\pdfoutput=1

\documentclass[11pt]{article}

\usepackage{emnlp2021}

\usepackage{times}
\usepackage{latexsym}

\usepackage[T1]{fontenc}

\usepackage[utf8]{inputenc}

\usepackage{microtype}

\usepackage[linesnumbered,ruled,vlined]{algorithm2e}
\SetKwInput{KwInput}{Input}                
\SetKwInput{KwOutput}{Output}              
\SetKwInput{KwReturn}{Return}              
\usepackage{algorithmic}
\usepackage{graphicx}
\usepackage{amsmath}
\usepackage{amssymb}
\usepackage{CJKutf8}
\usepackage{subcaption}
\usepackage{booktabs}
\usepackage{float}
\usepackage{multirow}

\newcommand{\baseline}{Offline }
\newcommand{\pseudoref}{Pseudo Refs }
\newcommand{\fixNAT}{Fixed + NAT }
\newcommand{\alignaware}{AlignAw }
\newcommand{\alignawareNAT}{AlignAw + NAT }
\newcommand{\alignawareNATAT}{AlignAw + NAT + AT }

\title{Monotonic Simultaneous Translation  with \\ Chunk-wise Reordering and Refinement}

\author{Hyojung Han$^1$\thanks{\quad Equal contribution} \thanks{\quad Work done at Samsung Research}  , Seokchan Ahn$^1$\footnotemark[1] \footnotemark[2], Yoonjung Choi$^1$, Insoo Chung$^1$\footnotemark[2], Sangha Kim$^1$,  Kyunghyun Cho$^2$ \\
$^1$Samsung Research, Seoul, Republic of Korea \\
 $^2$New York University, New York, United States \\
        hjhan@cs.umd.edu, seokchaa@uci.edu, yj0807.choi@samsung.com, \\ insoo.chung@tamu.edu, sangha01.kim@samsung.com, kyunghyun.cho@nyu.edu }

\begin{document}
\maketitle
\begin{abstract}

Recent work in simultaneous machine translation is often trained with conventional full sentence translation corpora, leading to either excessive latency or necessity to anticipate as-yet-unarrived words, when dealing with a language pair whose word orders significantly differ. This is unlike human simultaneous interpreters who produce largely monotonic translations at the expense of the grammaticality of a sentence being translated. In this paper, we thus propose an algorithm to reorder and refine the target side of a full sentence translation corpus, so that the words/phrases between the source and target sentences are aligned largely monotonically, using word alignment and non-autoregressive neural machine translation. We then train a widely used wait-$k$ simultaneous translation model on this reordered-and-refined corpus. The proposed approach improves BLEU scores and resulting translations exhibit enhanced monotonicity with source sentences.

\end{abstract}

\section{Introduction}
\label{sec:intro}
Simultaneous interpretation is widely used in various scenarios such as cross-lingual communication between international speakers, international summits, and streaming translation of a live video. 
Simultaneous interpretation has a latency advantage over conventional full-sentence translation, i.e. offline translation, as it requires only partial sequence to start translating.
However, as the source and target languages differ in word orders, there is a difficulty in simultaneous interpretation that does not exist in offline translation which translates only after the whole source sentence is received.
For example, when dealing with language pairs that significantly differ in word order (e.g., between SOV language and SVO language), an interpreter may not receive sufficient information with partial sequence to start generating a translation that respects the natural order of the target language. One of the approaches to address this problem is to perform \textit{anticipation}\footnote{A  simultaneous interpretation strategy where the interpreter says information that is not yet said by the speaker.}. 
Note that the nature of anticipation relies on interpreters' assumptions and the anticipation may provide incorrect translations.
Alternatively, human interpreters strategically resort to producing \textit{monotonic translations} that follow the word order of the source sequence \cite{zhongxi2020what}.

\begin{figure}[t]
    \centering
    \includegraphics[width=7.7cm,height=3.6cm]{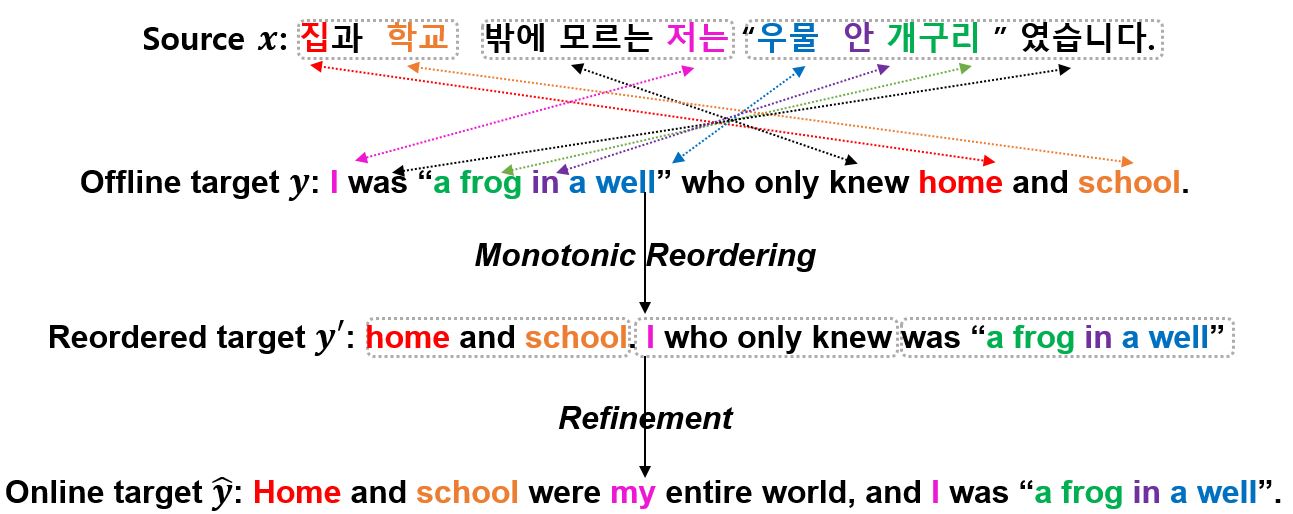} 
    \caption{An example illustration of monotonic reordering and refinement for simultaneous translation } 
    \label{fig:title_example}
\end{figure}

To illustrate the differences between the two strategies, the example in Figure \ref{fig:title_example} may be referred to. Of the two targets, offline target $\mathbf{y}$ respects the target language order and an online target $\mathbf{\hat{y}}$ roughly follows the source word ordering. Successful anticipation in Figure \ref{fig:title_example}'s case would be to predict the initial words in $\mathbf{y}$ (I was a ``frog in a well") before receiving the full $\mathbf{x}$. This would pose difficulty even to professional translators as all the relevant information is in the latter part of the $\mathbf{x}$ \begin{CJK}{UTF8}{} \CJKfamily{mj}(\small{저는$_{I}$ /``우물$_{a~well}$ / 안$_{in}$ / 개구리$_{a~frog}$" 였습니다$_{was}$.})\end{CJK}. \citet{anti2008} reports the success rate of human interpreters' anticipation attempts to be as low as 38.1\% even though they make predictions based on pre-acquired domain knowledge. On the other hand, a monotonic approach would be to generate an $\mathbf{\hat{y}}$ style translation - the grammaticality in the resulting sequence is sacrificed to translate only the received information.

A similar case applies to Simultaneous Machine Translation (SimulMT) models, which start translating before a whole sentence is given.
Several studies \cite{ma2018stacl, arivazhagan-etal-2019-monotonic, Ma2020Monotonic} often utilize offline full-sentence translation corpora to train SimulMT models. 
Offline full-sentence parallel corpora are expected to follow the natural order of languages and mostly contain source to offline-target pairs. Naturally, when SimulMT models are trained on these corpora, the models inevitably learn to perform anticipation. Recent SimulMT studies are focused on reducing anticipation \cite{zhang-etal-2020-learning-adaptive} or performing better anticipation \cite{zhang2020future}.
On the contrary, studies on enabling monotonic translation in SimulMT are scarcely available.  Recently, \citet{chen2020improving} suggest utilizing  pseudo-references for monotonic translation. 

In this paper, we propose a paraphrasing method to generate a monotonic parallel corpus to allow a monotonic interpretation strategy in SimulMT. Our method consists of two stages. The method first chunks source and target sequences into segments and monotonically \textit{reorders} the target segments based on source-target word alignment information (Section \ref{sec:chunk_reordering}). Then, the reordered targets are \textit{refined} to enhance fluency and syntactic correctness (Section \ref{sec:refine}). To show the effectiveness of our method, we train wait-$k$ models \cite{ma2018stacl} on the resulting monotonic parallel corpus of reordering-and-refinement. Results show improvements in BLEU scores over baselines and models producing monotonic translations. Our main contributions are as follows:

\begin{itemize}
\item We propose a method to reorder and refine the target side in an offline corpus to build a monotonically aligned parallel corpus for SimulMT.
\item We investigate the monotonicity in different language pairs, and show monotonicity can be improved after the reordering-and-refinement process.
\item We train widely used wait-$k$ models on generated monotonic parallel corpora in multiple language pairs. The results show improvements over baselines in both translation quality and monotonicity.
\end{itemize}

\begin{figure}[t]
    \centering
    \includegraphics[width=7.7cm,height=5cm]{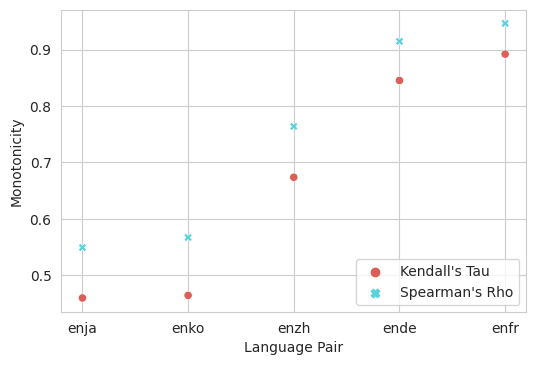}
    \caption{Monotonicity measured on offline trainsets. Utilized data is described in Section \ref{sec:datasets}. As Kendall's $\tau$ and Spearsman's $\rho$ show similar patterns, we only report Kendall's $\tau$ measurements in the rest of the paper.} 
    \label{fig:pair_wise_monotonicity}
\end{figure}

\section{Monotonicity Analysis}
\label{sec:monotonicity}

In this section, we analyze the degree of word order differences in multiple language pairs, i.e., the monotonicity in different language pairs. To measure the monotonicity, two rank correlation statistics are utilized: Kendall's $\tau$ and Spearman's $\rho$. The analyzed language pairs are: English-\{Korean, Japanese, Chinese, German, French\}.

According to \citet{polinsky2012head}, English is a head-initial language and Korean and Japanese are rigid head-final languages; Korean and Japanese are likely to exhibit extreme word order differences with English. German and Chinese are considered a mixture of head-final and head-initial languages; they are likely to have word differences with English, but not as severe as Korean or Japanese. French is also head-initial, so English and French pair is likely to have similar word order.

Figure \ref{fig:pair_wise_monotonicity} show monotonicity measurements between English and five different languages which vary in monotonicity: English-German and English-French pairs show high monotonicity, while English-Japanese and English-Korean pairs show low monotonicity. 

Lower monotonicity in language pairs presents higher difficulties for SimulMT tasks. 
For example, wait-$k$ algorithm only sees $k + t$ source tokens to generate a target token at step $t$ which could lead to unwanted anticipation. 
To avoid such anticipation, as we mentioned in Section \ref{sec:intro}, human interpreters often provide a monotonic translation. 
In the same sense, we conjecture that promoting monotonicity in training corpora is beneficial for translation quality in SimulMT.

\section{Monotonic Reordering and Refinement}
\label{sec:method}
In this section, we describe our proposed paraphrasing method of chunk-wise reordering and refinement to generate monotonic corpus for SimulMT.
Given source $\mathbf{x}=\{x_1, x_2, \cdots, x_{|\mathbf{x}|}\}$ and offline full sentence target $\mathbf{y}=\{y_1, y_2, \cdots, y_{|\mathbf{y}|}\}$, an alignment $\mathbf{a}$ is defined as a set of position pairs of $\mathbf{x}$ and $\mathbf{y}$.
$$\mathbf{a}=\{(s, t) : s \in \{1, \cdots, |\mathbf{x}|\}, t \in \{1, \cdots, |\mathbf{y}|\}\}$$
First, in chunk-wise reordering phase, we generate source chunk set $\mathcal{C}^{X}$ 
$$
\mathcal{C}^{X} = \{(x_{1:p_1}),(x_{p_1+1:p_2}), \cdots, (x_{p_{k-1}+1:p_k})\},\\
$$
where $0 < p_1 < p_2 < \cdots < p_k = |\mathbf{x}|$
and reordered target chunk set $\mathcal{C}^{Y}$ 
$$
\mathcal{C}^{Y} = \{(y'_{1:q_1}),(y'_{q_1+1:q_2}), \cdots, (y'_{q_{k-1}+1:q_k})\},\\
$$
where $0 < q_1 < q_2 < \cdots < q_k = |\mathbf{y}|$, and $y'_i \in \mathbf{y'}$ is reordered target token from offline target $\mathbf{y}$.
The elements of a reordered target chunk $\mathcal{C}^{Y}_i$ are corresponding target tokens of a source chunk $\mathcal{C}^{X}_i$ based on given alignment information $\mathbf{a}$. Also, we preserve the original target order within each $\mathcal{C}^{Y}_i$. For example, offline and reordered target in Figure \ref{fig:title_example}  correspond to $\mathbf{y}$ and $\mathbf{y'}$ respectively, and both sequences are only different in token orders.
The number of source chunks in one sentence is the same as the number of reordered target chunks ($|\mathcal{C}^{X}|=|\mathcal{C}^{Y}|$), while the number of tokens in $|\mathcal{C}^{X}_i|$ and $|\mathcal{C}^{Y}_i|$ could vary.
We experiment two chunking methods; fixed-size chunking and alignment-aware adaptive size chunking. 

Given chunked sets $\mathcal{C}^{X}$ and $\mathcal{C}^{Y}$, we refine reordered target tokens to generate more natural and fluent sentence with a Non-Autoregressive Translation (NAT) model. In the refinement algorithm, final paraphrased sentence $\mathbf{\hat{y}}$ is generated from reordered sequence $\mathbf{y'}$. Furthermore, we incorporate an Autoregressive Translation (AT) model into our refinement process.
The more detailed steps for each phase will be explained in the following subsections.

\subsection{Chunk-wise Reordering}
\label{sec:chunk_reordering}
\subsubsection{Fixed-size Chunk Reordering}
\label{sec:fixed_chunk_reordering_chunk}
In the fixed-size chunk reordering method, we simply chunk a sequence of tokens into fixed size segments. 
The source chunk set $\mathcal{C}^{X}$ in this chunking method is as follows:
$$\mathcal{C}^{X} = \{(x_{1:K}),(x_{K+1:2K}), \cdots, (x_{[|\mathbf{x}|/K]:|\mathbf{x}|})\},$$
where $K \in [1, |\mathbf{x}|]$ is chunk size. If $k=1$, $\mathcal{C}^{X}$ is identical with $\mathbf{x}$. 
We conduct subword operation such as sentencepiece or BPE after chunking process in order to avoid subword separation.

\begin{figure*}[t]
    \centering
    \includegraphics[width=16cm,height=5cm]{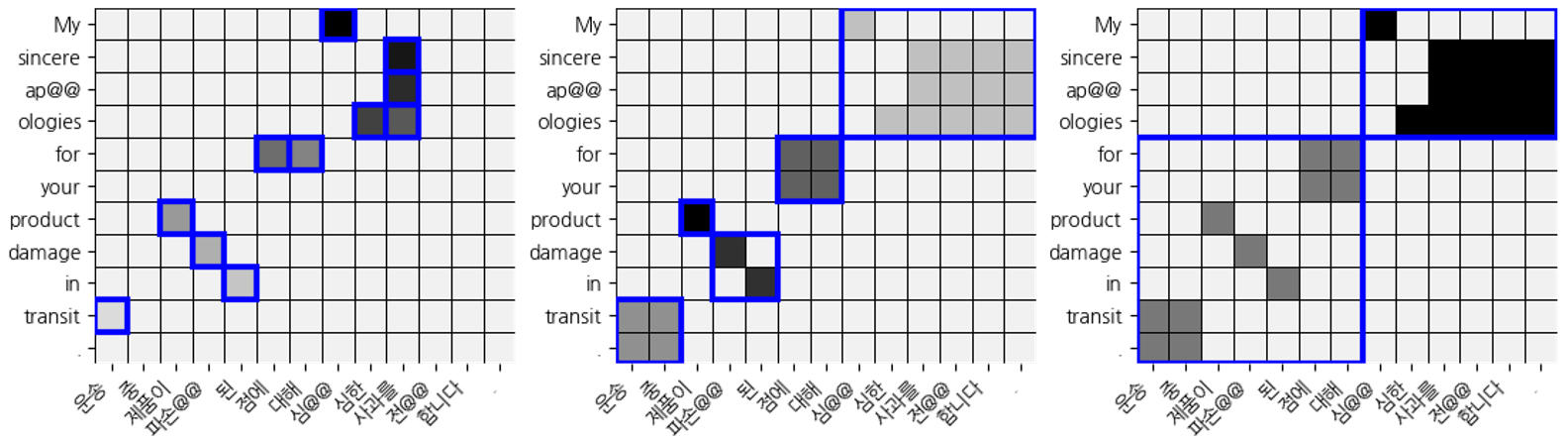}
    \caption{Example of alignment-aware reordering process. \textbf{Left}: Original alignments. \textbf{Center}: (Initialization) After filling align vacancy, merging mid-split subwords and enforcing the consistency requirement.  \textbf{Right}: After merging all the chunks shorter than length thresholds. } 
    \label{fig:consist_merging_example}
\end{figure*}

\subsubsection{Alignment-Aware Chunk Reordering}
\label{sec:alignment_aware_chunk_reordering}

In the alignment-aware chunking method, we segment a sentence adaptively by leveraging alignment information $\mathbf{a}$, as described in Algorithm \ref{algo:reorder}.
The left grid in Figure \ref{fig:consist_merging_example} presents the subword alignment between source and target sentence. We run aligner on subword over word because the alignment performance is consistently better \citet{zenkel-etal-2020-end} when using GIZA++ \cite{och03:asc} , which we use in our experiments. Based on this alignment information, we initialize a list of chunks $\mathcal{C}$. As observable, there are some tokens which have no alignment information. To avoid omission, we assign the same alignment as the previous token; if a token is at the head, it follows the next token's alignment. 
To ensure subwords can be properly detokenized after reordering, we merge  mid-splitted subwords.
The middle grid in Figure \ref{fig:consist_merging_example} presents the result of these initialization steps.
After initialization, we generate consistent chunks by merging all the inconsistent ones, following the definition of consistency in \citet{10.1007/3-540-45751-8_2}. In a consistent chunk, tokens are only aligned to each other, not to tokens in other chunks. If any chunk in $\mathcal{C}$ has size smaller than a minimum size threshold $\delta$, we merge a chunk pair that are adjacent in both source and target side and have the shortest target distance between them.
If the distances are the same between multiple candidate pairs, we choose the pair of chunks that makes the smallest size after merging. 
We additionally merge the chunks adjacent to the merged one if they are arranged monotonically. Merging is repeated until all chunks meet the size requirements. An example of final result is the right grid in Figure \ref{fig:consist_merging_example}. Phrase extraction method used in statistical machine translation \citet{10.1007/978-3-540-30194-3_13} also makes phrase level alignments from word alignments using heuristics like ours, but it tends to choose shorter phrases since the number of co-occurrences decrease drastically as the phrase size grows, which makes it difficult to generate larger chunks to prevent hurting grammatical correctness while reordering phase.

\begin{algorithm}[]
    \SetKwFunction{Aligner}{Aligner}
    \SetKwFunction{FillAlign}{FillAlign}
    \SetKwFunction{MergeSubword}{MergeSubword}
    \SetKwFunction{MergeInconsistent}{MergeInconsistent}
    \SetKwFunction{SelectClosestPair}{SelectClosestPair}
    \SetKwFunction{MergeAdjacent}{MergeAdjacent}
    \SetKwFunction{MergeMonotonic}{MergeMonotonic}
    \SetKwFunction{ReorderTarget}{ReorderTarget}
    
    \DontPrintSemicolon
    \SetAlgoLined
    \KwInput{Source sentence $\mathbf{x}$ and target sentence $\mathbf{y}$}
    \KwOutput{Monotonically aligned chunks $\mathcal{C'}$}
    $\mathbf{a} =$ Alignment between $\mathbf{x}$ and $\mathbf{y}$\;
    $\mathcal{C} =$ Initialize chunks $\mathcal{C}$\;
    $\mathcal{C} =$ Merge all the inconsistent chunks in $\mathcal{C}$ \;
    
    \While{
        $|C_{i}^{X}| < \delta_{src}\ \mathbf{or}\ |{C}_{i}^{Y}| < {\delta}_{tgt}$
        $\mathbf{for\ any}\ \mathcal{C}_{i}\ \mathbf{in}\ \mathcal{C}$
     }
     {
      $\mathcal{C}_{k} =$ The smaller of the chunks adjacent to $\mathcal{C}_{i}$\;
      Merge $\mathcal{C}_{i}$ and $\mathcal{C}_{k}$\;
      Merge monotonic chunks adjacent to $\mathcal{C}_{i}$\;
     }
     $\mathcal{C'} =$ Reorder target side of $\mathcal{C}$ monotonically\;
     
     \caption{Alignment-Aware reordering}
     \label{algo:reorder}
\end{algorithm}

\subsection{Refinement}
\label{sec:refine}
Reordered target results from previous phase inevitably entail irregularities mainly for two reasons. One could be broken connectivity of collocations in segmentation process. 
The other would be disfluently missing or containing words of endings and preposition as the position of chunk has been changed, thus requiring an addition of new words or clearing unnecessary words.
In this part, we focus on refining aforementioned anormalities in order to enhance fluency, while preserving the monotonicity at the same time.

\subsubsection{Refinement with NAT}
\label{sec:refine_NAT}
\begin{algorithm}
\DontPrintSemicolon
\SetAlgoLined
\KwInput{Source and target chunks $\mathcal{C}^{X}, \mathcal{C}^{Y}$}
\KwOutput{Paraphrased target $\mathbf{\hat{y}}$}
$i = 1$\;
$\hat{Y}_0 = []$
\;
 \While{$i \leq |\mathcal{C}^{X}|$}{
  $X_i  = \mathcal{C}^{X}_{1:i}$ and
  $Y'_i = [\hat{Y}_{i-1}; \mathcal{C}^{Y}_i]$ \;
$  \hat{Y}_i  = \arg\max_{Y}\log p^\mathcal{R}(Y|X_i, Y'_i) $ \;
$i = i + 1$\;
 } 
\KwReturn{ $\hat{Y}_{|\mathcal{C}^{X}|} $}
 \caption{Chunk-wise Refinement}
 \label{algo:refine}
\end{algorithm}

We iteratively decode partial source $\mathcal{C}^{X}$ with pre-trained translation model, given partial reordered target $\mathcal{C}^{Y}$ as a guidance in order to generate corresponding online target $\hat{Y}$. 
More specific process is explained in Algorithm \ref{algo:refine}.
As the model refines given $[\hat{Y}_{i-1}; \mathcal{C}^{Y}_i]$, previous refined output $\hat{Y}_{i-1}$ could be altered as the model re-generates the entire sequence from scratch.
Similarly in re-translation \cite{re-translation, han2020faster}, we set an option of fixed or alterable prefix to force the model whether to generate same target prefix of $\hat{Y}_{i-1}$ or to allow the model to modify the prefix. 
As we limit the visibility of source information and iteratively generate target tokens with increasing source chunks, we expect the refinement model to generate monotonically aligned and paraphrased targets $\mathbf{\hat{y}}$ with enhanced fluency.

We use NAT architecture as the core refinement model $\mathcal{R}$.
In NAT inference, the model's decoder is first given source features and fed an empty target sequence. Then the NAT decoder develops the empty sequence into a translation of the source sequence. This development is often iterative. Note that at every iteration step, the target sequence is refined - closer to the source sequence in meaning and become more fluent.
This motivates us to utilize NAT architecture in our refinement process for monotonic-yet-disfluent sequences.
In our approach, the NAT model starts refinement iteration with initialized tokens of previous output and reordered target chunk $ Y'_i$, instead of an empty sequence. 
This target initialization act as a weak supervision to generate monotonically aligned target, which allow model to focus only on the fluency the reordered targets. 

\subsubsection{Incorporation of AT }
\label{sec:refine_NAT_AT}
Despite the aptness of NAT structure to our refinement phase, 
NAT model entails a performance degradation compared to AT model in the expense of speedup.
Also, there exists repetition problem in NAT \citep{lee-etal-2018-deterministic, gu2020fully} which is generated in the process of multiple chunk-wise iterative refinement.
In order to complement the aforementioned weaknesses of NAT decoding, we incorporate AT into our refinement process with NAT model.
The final probability is computed jointly with the probability of AT and NAT model:
\begin{multline}
p^\mathcal{R}(Y|X_i, Y'_i)  \propto \\
p^{AT}(Y | X_i)^{\alpha} \cdot p^{NAT}(Y |X_i, Y'_i)^{(1-\alpha)},
\label{eq:joint_prob}
\end{multline}
where $\alpha \in [0, 1]$ is hyper-parameter deciding the ratio between AT and NAT probability. 


\begin{table}[H]
\centering

\begin{tabular}{ccccc}
\toprule
\small Size/$L_{avg}$          & \small EnKo & \small EnJa & \small DeEn & \small EnZh  \\
\hline
\small Train                    & \small3.4M/22 & \small3.9M/12 & \small4M/28   & \small15.9M/27 \\
\small Valid                    & \small800/19  & \small4451/17 & \small3000/25 & \small4000/30  \\
\small Test                     & \small1429/21 & \small1194/17 & \small2169/25 & \small4000/30  \\
\small AlignAw                  & \small3.1M & \small1.7M & \small2.2M & \small6.5M \\ 
$er$                     & \small 0.85 & \small 1.15 & \small 0.98 & \small 1.12  \\ \bottomrule


\end{tabular}

\caption{Data statistics and average of En token length $L_{avg}$ of used corpus. AlignAw denotes number of pairs processed with alignment aware refinement. $er$ denotes emission rates used in wait-$k$ decoding. Token lengths and $er$ are measured base on subwords counts. }
\label{table:data_statistics}
\end{table}



\section{Experiments}
\subsection{Dataset}
\label{sec:datasets}

In this section, we describe the utilized datasets. Detailed statistics are presented in Table \ref{table:data_statistics}. Utilized EnKo trainset and devset are created using in-house translation corpora while test scores are reported on IWSLT17 \cite{iwslt} EnKo testset. The DeEn trainset of WMT15 translation task \cite{bojar-etal-2015-findings} is utilized. \textit{newstest2013} is utilized as devset and \textit{newstest2015} is used as testset. The EnJa trainset and validset are respectively the combination trainsets and validsets of KFTT \cite{neubig11kftt}, JESC \cite{pryzant_jesc_2018}, TED \cite{wit3}. The trainset and validset are used as preprocessed and provided by the MTNT authors\footnote{\url{https://www.cs.cmu.edu/~pmichel1/mtnt/}} \cite{mtnt}. Only the TED portion of testsets is used. For EnZh training UN Corpus v1.0 \cite{uncorpus} is used. Trainset, devset, and testset follow the original splits. Monotonicity of EnFr in Figure \ref{fig:pair_wise_monotonicity} is measured on the WMT14 \cite{bojar-etal-2014-findings} trainset. Additional details regarding utilized tokenization and vocabulary training are listed in Appendix \ref{app:dataset-details}.

\subsection{Metric} All the BLEU scores are cased-BLEU measured using sacreBLEU \citep{post-2018-call}. Test scores are measured using models that report best BLEU on their respective devsets.
All references and translations of each Korean, Japanese, and Chinese languages are tokenized prior to BLEU evaluation. Tokenizers utilized are mecab-ko\footnote{\url{https://github.com/hephaex/mecab-ko}}, KyTea \footnote{\url{http://www.phontron.com/kytea/}},
and jieba for Korean, Japanese and Chinese respectively. We report detokenized BLEU on DeEn results. To measure monotoniticy, we use Kendal's $\tau$ rank correlation coefficient.

\subsection{Implementation Details}

The default setting for NMT and SimulMT models follow the base configuration of transformer \citep{vaswani2017attention}. SimulMT models are trained using wait-$k$ algorithm, where $k \in \{4, 6, 8, 10, 12\}$, with uni-directional encoder similarly to \citet{han-etal-2020-end}. The base NMT and SimulMT models are trained up to 300k train steps on a single GPU - each step is performed on a batch of approximately 12288 tokens. For refinement, we utilize NAT models of Levenshtein transformer architecture \cite{gu2019levenshtein} with maximum iteration of 1. The NAT models are trained using sequence-level knowledge distillation \cite{kim-rush-2016-sequence} - the references of trainset pairs are replaced with beam search results ($beam=5$) of NMT teachers. The NAT models follow base configurations and teacher NMT models follow big configuration. Both types of models are trained up to 300k steps on 8 GPUs. In each training step, a 8192 tokens batch is used per GPU. Additional implementation details can be found in the Appendix \ref{app:implementation-details}.
\begin{table*}[]
\centering
\begin{tabular}{lccccc}
\toprule
Wait-$k$ BLEU                                         & $k$=4       & $k$=6       & $k$=8       & $k$=10      & $k$=12      \\
\hline
\baseline                                    & 10.9    & 12.1    & 12.6    & 12.8    & 13.4    \\
\fixNAT                                    & 11.2    & 12.3    & \textbf{13.5}    & \textbf{13.5}    & \textbf{13.7}    \\
 \alignawareNAT                                & 11.4    & \textbf{12.8}    & 12.7    & 12.9    & 13.1    \\
 \alignawareNATAT  ($\alpha=0.25$) & \textbf{11.7}    & 12.6    & 12.4    & 12.7    & 13.2    \\
 \alignawareNATAT  ($\alpha=0.50$) & 10.7    & 12.4    & 13.0      & 13.0      & 13.2    \\
\toprule
Wait-$k$ Kendal's $\tau$                                 & $k$=4       & $k$=6       & $k$=8       & $k$=10      & $k$=12      \\
\hline
\baseline                                     & 0.65526 & 0.60792 & 0.58440 & 0.55169 & 0.53701 \\
 \fixNAT                                      & 0.71822 & 0.66811 & 0.63154 & 0.61244 & 0.59478 \\
 \alignawareNAT                                & 0.71903 & 0.69101 & 0.67149 & 0.65985 & 0.64043 \\
 \alignawareNATAT  ($\alpha=0.25$) & \textbf{0.73215} & 0.69386 & 0.69524 & 0.67593 & 0.63335 \\
 \alignawareNATAT  ($\alpha=0.50$) & 0.73055 & 0.70106 & 0.67674 & 0.65559 & 0.64042 \\
\bottomrule
\end{tabular}

\caption{BLEU scores and monotonicity measurements of EnKo wait-$k$ models trained on offline and reordered-and-refined corpora. Note that monotonicity is measured between the model translations and testset references.}
\label{table:koen_para_vari}
\end{table*}

\subsection{Corpus Generation and Training}
We demonstrate the effectiveness of our reordering-and-refinement method by training wait-$k$ models on the resulting datasets. The wait-$k$ models are trained on the combination of the monotonically aligned training pairs and offline trainset. 
\alignawareNATAT denotes monotonically aligned corpora using alignment-aware reordering and refinement using joint probability of NAT and AT models. 
And \baseline refers to the offline full-sentence corpus. 

\subsubsection{Reordering}
\paragraph{Fixed:} For fixed-size reordering, we experiment with chunk sizes $K \in \{4, 6, 8, 10, 12\}$. 
In wait-$k$ training, $k$ and $K$ are matched. 
All fixed-size reordered-and-refined corpora have the same size as corresponding offline corpus.
\paragraph{AlignAw:} For each corpus, we generate four variations of alignment-aware reordering with source and target minimum chunk size of ${2, 3}$. Alignment-aware reordering is not applicable on the already-monotonic cases and the sentence pairs which are locally non-monotonic inside a chunk and globally monotonic among chunks within a single pair - typically, the reordering method is applicable to 20\% to 50\% of offline corpus.

We gather unique pairs from the created four variations to generate the final reordered pairs. The statistics of reordered set for each translation direction is in Table \ref{table:data_statistics}. The resulting pairs are refined and combined with corresponding offline corpus to train wait-$k$ models. Here, same set of reordered pairs are utilized for all $k$ settings.

\subsubsection{Refinement}
\paragraph{NAT:} NAT models are utilized to refine the reordered pairs. Both the fixed prefix and alterable prefix refinement is performed and combined. BertScore \citep{bert-score} is measured and used to discard refinement results that show below average scores. The size of the resulting set is the same as the corresponding offline corpus.
\paragraph{NAT + AT:} 
NAT and AT models can both be utilized to jointly compute token probability in refinement (Section \ref{sec:refine_NAT_AT}). The AT models utilized are the baseline wait-$k$ models trained on offline corpora. We experiment with $\alpha \in \{0.25, 0.5\}$. The examples of reordered-and-refined sequences can be found in Appendix \ref{app:examples}.

\begin{table}[]
\centering

\begin{tabular}{lcccc}
\toprule
seq-rep-$n$       & \baseline & NAT & NAT + AT    \\
       \midrule
1-gram & 0.036   & 0.076         & 0.072 \\
2-gram & 0.008   & 0.022         & 0.018 \\
3-gram & 0.003   & 0.009         & 0.006 \\
4-gram & 0.001   & 0.004         & 0.002 \\
\bottomrule
\end{tabular}

\caption{N-gram repetition rate measured on offline and reordered-and-refined EnKo corpora. $\alpha=0.25$ is set to for NAT + AT}
\label{table:repetition}
\end{table}

\begin{figure*}[ht]
    \centering
    \includegraphics[width=16cm,height=4cm]{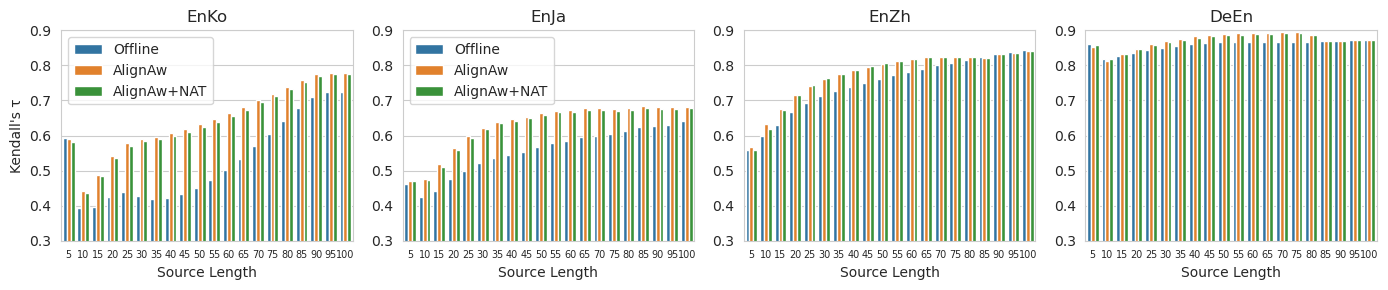}
    \caption{Monotonicity of offline pairs and pairs processed with reordering-and-refinement. AlignAw indicate that targets are alignment-aware reordering (Section \ref{sec:alignment_aware_chunk_reordering}, and \alignawareNAT show monotonicity after NAT refinement (Section \ref{sec:refine_NAT}) is applied to AlignAw pairs.) }
    \label{fig:monotonicity_modification_orig_reorder}
\end{figure*}
 \begin{figure*}[h] 
  \centering
   \begin{subfigure}[t]{1.0\textwidth}
    \centering
     \includegraphics[width=1.0\textwidth]{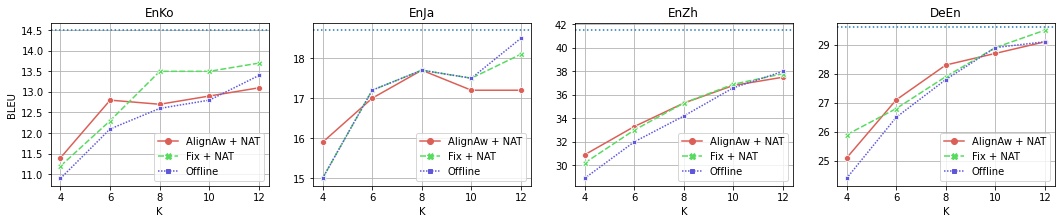}
    \end{subfigure}
    \hspace{-1cm}
    \begin{subfigure}[t]{1.\textwidth}
        \centering
        \includegraphics[width=1.0\textwidth]{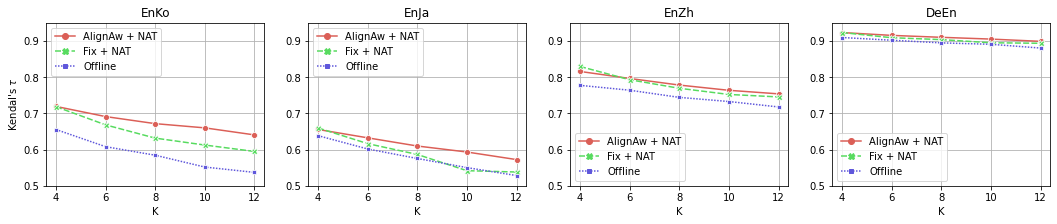}
    \end{subfigure}
    \caption{
    BLEU scores and monotonicity measurements presented by wait-$k$ models trained on offline translation corpora and variations of reordered-and-refined corpora.}
    \label{fig:BLEU_Kendal_K_pairs}%
\end{figure*}

\section{Results and Analysis}
\subsection{Experimental Results on EnKo}
Table \ref{table:koen_para_vari} shows BLEU scores and Kendal's $\tau$s of wait-$k$ models trained using original offline corpus and variations of reordered-and-refined corpus.
We observe that the models trained on monotonically reordered-and-refined corpora show higher BLEU scores and monotonicity. 
\paragraph{Reordering:} Of the variations, corpora including AlignAw chunking process generally show better BLEU scores over \fixNAT when $k \leq 6$. This could be the benefit of the semantically plausible way to split sentences provided by AlignAw chunking. On the other hand, models trained with \fixNAT corpora show higher BLEU when $k \geq 8$.
\paragraph{Refinement:} 
Experiments on utilizing AT probabilities show degraded BLEU scores in $k \in {6,8,10}$.
On the contrary, the models trained on \alignawareNATAT corpora show  enhanced monotonicity. 
The $\alpha$ value may be adjusted to make trade-off  between promoting monotonicity in translation or enhancing translation quality in terms of BLEU.


\paragraph{Repetition Reduction with AT:} 
Following \citep{Welleck2020Neural}, we report $n$-gram repetition rate, seq-rep-$n$, on each generated corpus in Table \ref{table:repetition}. 
We observe from seq-rep-$n$ in all of the tested $n$ values, that employing AT models in refinement help alleviating the repetition problem of posed by NAT models. 

\subsection{Language Pairs Comparison}
Figure \ref{fig:monotonicity_modification_orig_reorder} shows the difference in monotonicity between different language pairs: EnKo, EnJa, EnZh, and DeEn. 
It is observable in Figure \ref{fig:monotonicity_modification_orig_reorder} that the overall monotonicity in EnKo and EnJa pairs is enhanced after paraphrasing, while monotonicity scores of DeEn remain almost the same, only showing slight improvement. The extent of monotonicity enhancement in EnZh is between that of EnKo/Ja and DeEn. 
In all language pairs, the enhancements are generally lower in long or very short sequences.
In the case of long sequence pairs, a pair may contain multiple sequences and be already aligned at the sequence level, thus resulting in marginal monotonicity enhancement.
In the case of shorter length sequences, the whole sentence may be merged into a single chunk, less benefiting from our process. 
After the reordered sets are refined, monotonicity marginally decreases. This is expected as forcibly aligned tokens are refined to augment the fluency in the resulting sentence.

To present the effectiveness of generated monotonic corpus in different language pairs, we train wait-$k$ models on EnKo, EnJa, EnZh, and DeEn, and report BLEU and Kendal's $\tau$ of the models in Figure \ref{fig:BLEU_Kendal_K_pairs}. The horizontal dotted line presents the BLEU of unidirectional offline model.
The important observation we can find is that the monotonicity increment of wait-$k$ model in Figure \ref{fig:BLEU_Kendal_K_pairs} is proportional to that of generated monotonic corpus in Figure \ref{fig:monotonicity_modification_orig_reorder}, suggesting that promoting monotonicity in training corpus is beneficial for SimulMT models to generate monotonic output, especially in language pairs with differing word orders.

Within two paraphrasing methods, \fixNAT and \alignawareNAT, we can see that the monotonicity of \fixNAT is always in between that of \baseline and \alignawareNAT   in Figure \ref{fig:BLEU_Kendal_K_pairs}, and the gap increases as the $k$ value get higher. 

While our methods are effective in EnKo, the performance of suggested method is similar or lower than that of baseline in EnJa.
We presume that the ineffectiveness in EnJa is due to its short average sentence length with highest emission rate, as shown in Table \ref{table:data_statistics}.
A short sentence often cannot preserve semantic properties while being split into chunks and reordered. For example, Fixed-length reordering always chunks all sentences ignores such feature and only increases disfluency in the chunked result.
Also, even though \alignaware enforces the consistency requirement on the chunks, adjustment of such requirement like changing minimum chunk size may be required considering the high emission rate.

Based on the highest performance at $k=8$, there is about 8\% BLEU improvement over \baseline in EnKo whereas there is about 3\% improvement in EnZh and about 2\% improvement in DeEn. It is roughly proportional to the monotonicity improvements shown in Figure \ref{fig:monotonicity_modification_orig_reorder}.

\subsection{Evaluation on Online References}
\begin{figure}[t]
    \centering
    \includegraphics[width=7.7cm,height=5.0cm]{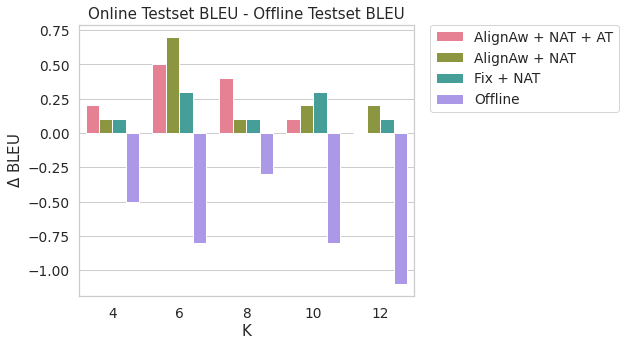}
    \caption{Differences of BLEU score between online and offline of EnKo wait-$k$ models.}
    \label{fig:on_vs_off}
\end{figure}

We test wait-$k$ models on our in-house EnKo online and offline testsets of 150 lines. We choose EnKo because the impact of reordering-and-refinement is the greatest in that pair. The source sentences of both testsets are identical. The online references are constructed by a professional interpreter under a simulated simultaneous interpretation scenario and the offline references are constructed by the same interpreter assuming a typical translation scenario. In construction of online references interpreter was encouraged to perform monotonic interpretation rather than anticipation.
BLEU scores are computed with both online and offline references for each trained model.
Figure \ref{fig:on_vs_off} plot the subtraction of BLEU scores on offline references from BLEU scores on online references.
It is noticeable that the wait-$k$ models trained on offline corpus have negative value while all the models trained on generated corpus present positive values, which implies the effectiveness of our approach. 
Overall, the substantial differences at $k=6$ may suggest that the chunk size utilized by human interpreter has comparable value. 

\section{Related Work}

Due to word order differences between languages, SimulMT training often face situations where anticipation is required.
Note that word order difference is observed to be problematic even for human interpreters \cite{alya2004, tohyama-matsubara-2006-collection}.
\citet{chen2020improving} suggest using pseudo-references which involve utilizing wait-$k$ inference output to limit "future anticipation" in training.
\citet{zhang2020future} utilize NMT teachers to implicitly embed future information in their SimulMT students for better anticipation performance. \citet{zhang-etal-2020-learning-adaptive} study adaptive policy to tackle this problem - 
authors suggest an adaptive SimulMT policy that dictate READ/WRITE actions based on whether "meaningful units" are fully formed with consumed input tokens.

Related work in the broader SimulMT and paraphrasing domain is presented in Appendix \ref{app:related-work}.

\section{Conclusion}
Most of SimulMT models are trained on offline translation corpora, which could lead to limitation in translation quality and achievable latency, especially in non-monotonic language pairs.
To address this problem, we propose a reordering-and-refinement algorithm to generate monotonically aligned online target with NAT model. 
We then train widely used wait-$k$ SimulMT models on this newly generated corpus. 
Resulting models show BLEU score improvement and significant enhancement on monotonicity in multiple language pairs.
\bibliography{anthology}
\bibliographystyle{acl_natbib}

\clearpage

\appendix

\label{sec:appendix}

\section{Dataset Details}
\label{app:dataset-details}

All utilized texts regarding English-to-Korean and German-to-English directions are first tokenized with Moses \cite{moses}, then per-language BPE vocabularies are learned on the moses-tokenized trainset. The sizes of the vocabularies are 29k BPE English vocabulary and 44k BPE Korean vocabulary for English-to-Korean and 16k BPE German vocabulary and 16k BPE English vocabulary for German-to-English. The English-to-Japanese texts are first moses-tokenized. And KyTea\footnote{\url{http://www.phontron.com/kytea}} is applied to additionally tokenize Japanese texts. Separate English and Japanese vocabulary of size 32K is trained on tokenized training data using Sentencepiece \cite{kudo-richardson-2018-sentencepiece}. For English-to-Chinese training, no moses-tokenization is applied and Chinese sentences are tokenized using Jieba\footnote{\url{https://github.com/fxsjy/jieba}}. Separate English and Chinese vocabulary of size 32K is trained on training pairs using sentencepiece. The in-house EnKo data consists mainly of the AIHub EnKo offline translation corpus\footnote{\url{https://aihub.or.kr}}, news domain translation data, in-house proprietary patent data, and translated dialogue data of general domain.

\section{Additional Implementation Details}
\label{app:implementation-details}
Our implementation is based on fairseq \cite{ott2019fairseq}, and all GPUs used are V100s. The alignment information used in reordering process is extracted with GIZA++ \cite{och03:asc}. The alignment information used to evaluate monotonicity is extracted using fast-align \cite{dyer-etal-2013-simple}. The alignments are measured in subword level. Embedding weights are separately learned for source and target languages, while transposed target language embedding weights also works as linear projection layers at the top of transformer decoders.

\section{Reporting AlignAw + NAT Scores}
\label{app:choosing-aa}
The \alignawareNAT wait-$k$ models are trained on different variations of \alignawareNAT corpora - \alignawareNAT corpora generated with prefixes fixed ($\mathbf{b0}$), and with alterable prefixes ($\mathbf{b1}$), and combination of $\mathbf{b0}$ and $\mathbf{b1}$ filtered using BertScore ($\mathbf{b0b1}$). The reported \alignawareNAT testset BLEU scores are of the wait-$k$ models that show highest BLEU score on validset regardless of the dataset variations.

\section{Comparison with Test Time wait-k Refs}
In recent work, \citet{chen2020improving} propose a method of pseudo-references generated with test time wait-$k$ decoding.
We apply their method in EnKo to create pseudo references for $k \in \{4, 6, 8, 10\}$ and train wait-$k$ model.
The results are presented in Table \ref{table:pseudorefs}.
Similar to our monotonicity metric, this work also suggest $k$-anticipation rate ($k$-AR) as a metric of parallel corpora. 
We also measure and report our generated corpus with this metric.
Compare to \baseline and \pseudoref, we see that our \alignawareNAT corpus significantly decrease $k$-AR and the models trained with \alignawareNAT also show enhanced BLEU score in general.

\begin{table}[]
\centering
\begin{tabular}{lcccc}
\toprule
\small EnKo  Wait-$k$      & $k$=4    & $k$=6    & $k$=8    & $k$=10   \\ \hline
\small\baseline      &  \small10.9/18 &  \small12.1/12 &  \small12.6/7 &  \small12.8/4 \\
\small\pseudoref &  \small11.2/19 &  \small11.6/13 &  \small12.2/9 &  \small13.3/5 \\
\small\fixNAT    &  \small11.2/12 &  \small12.3/8 &  \small13.5/5 &  \small13.7/3 \\
\small\alignawareNAT   &  \small11.4/9 &  \small12.8/6 &  \small12.7/3 &  \small12.9/2 \\ 

\bottomrule
\end{tabular}
\caption{BLEU scores/$k$-AR\% of EnKo wait-$k$ models.}
\label{table:pseudorefs}
\end{table}

\section{Examples of Paraphrased Targets}
\label{app:examples}

\begin{figure*}[t] 
  \centering
    \centering
    \includegraphics[width=1.0\textwidth]{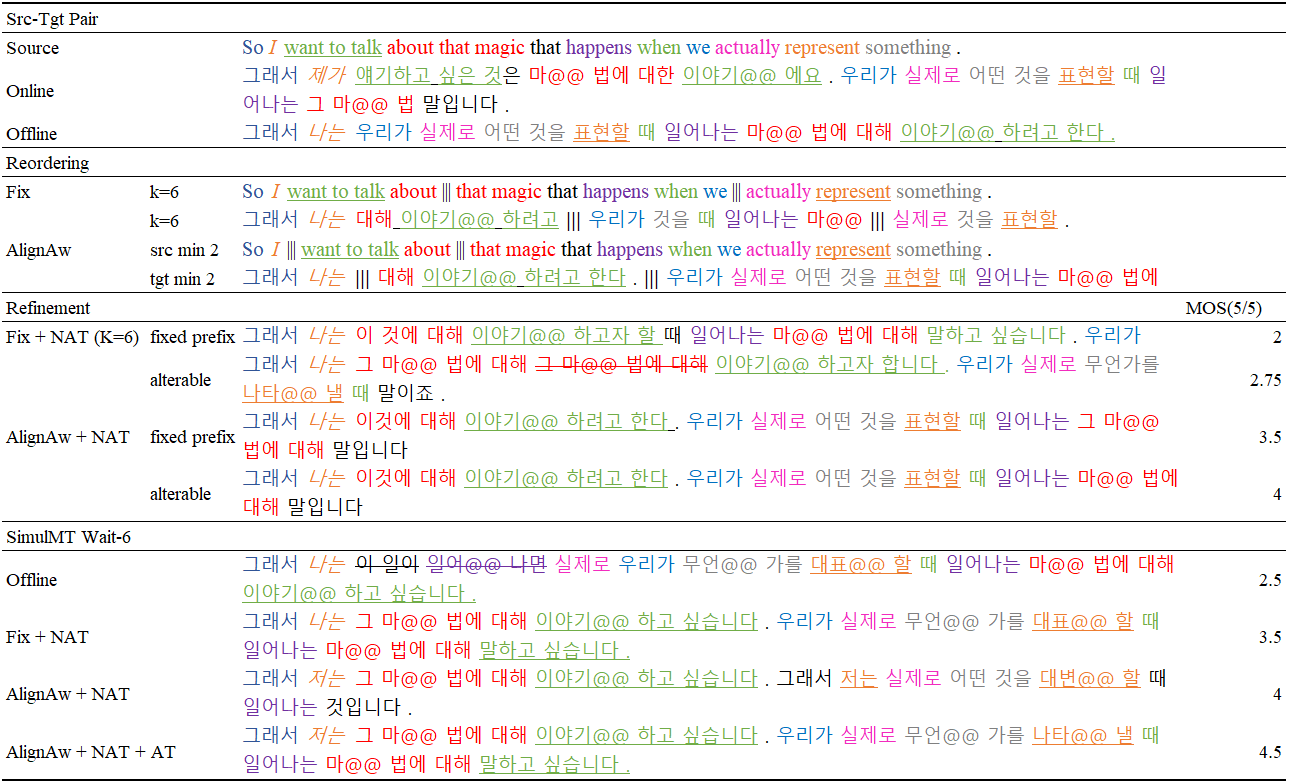}
    \caption{Example sentence of EnKo in whole pipeline process from inputs to SimulMT results.  }
    \label{fig:ex_ko_text}
 \end{figure*}

\begin{figure*}[t]    
    \centering
    \includegraphics[width=0.5\textwidth]{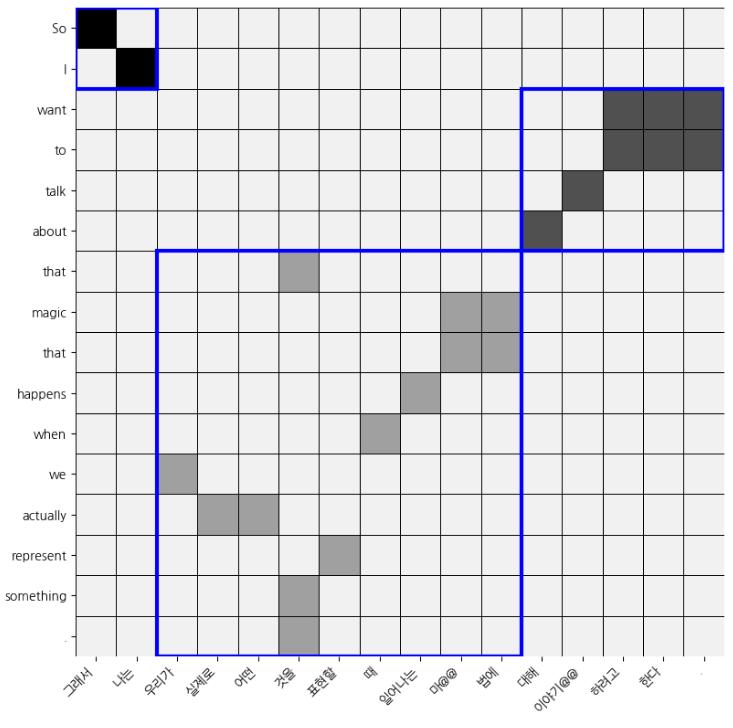}
       
    \caption{
    Alignment-aware result of the EnKo pipeline example in Figure \ref{fig:ex_ko_text}}
    \label{fig:ex_ko_alignaw}
\end{figure*}

\begin{figure*}[t] 
         \centering
        \includegraphics[width=1.0\textwidth]{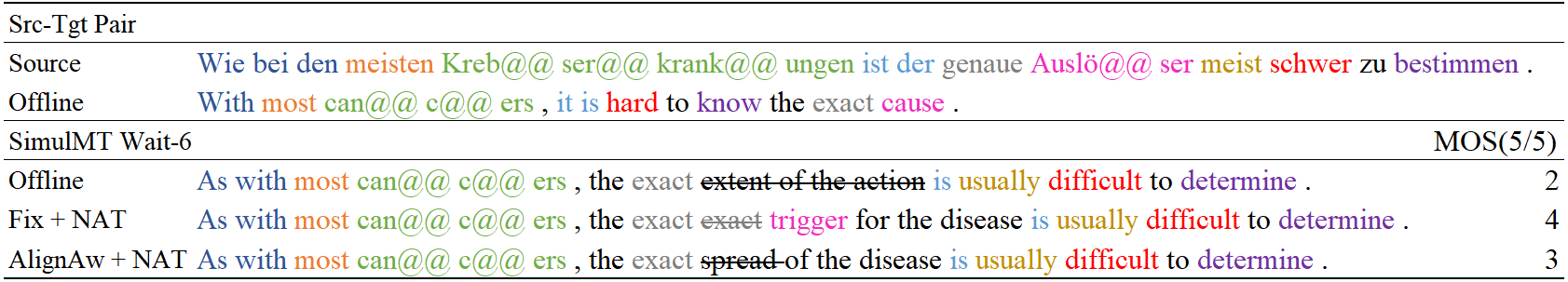}
        
    \caption{
    Example outputs of DeEn wait-$k$ models trained on reordered-and-refined corpora}
    \label{fig:ex_de_waitk}
\end{figure*}

Figure \ref{fig:ex_ko_text} presents an example sentence of English to Korean in whole pipeline process. 
We first represent the source sentence and its two different target sentences, online and offline translation. 
As results of the reordering phase, for each method (i.e., fixed chunking and AlignAw chunking), we provide only one case: $k=8$ in fixed chunking and $\delta_{src}=2$ and $\delta_{tgt}=2$ in AlignAw chunking.
Figure \ref{fig:ex_ko_alignaw} shows the final grid of AlignAw chunking in this example. 
We conduct the MOS evaluation with the result of refinement phase. MOS is the average of human-evaluated score by professional interpreters. In this evaluation, AlignAW + NAT shows the best performance than others.
Moreover, we present the inference outputs of SimulMT models which are trained on generated monotonic corpus. In this case, results of our methods are better than the result of offline model. 
We also provide DeEn example in Figure \ref{fig:ex_de_waitk}.

\section{More Related Work}
\label{app:related-work}

\paragraph{Simultaneous Translation:}
A fixed policy is used in \cite{dalvi-etal-2018-incremental} and \cite{ma2018stacl} which train SimulMT models according to the pre-defined policy. In particular, the Wait-$k$ strategy proposed by \cite{ma2018stacl} waits for $k$ sub-words and alternates READ/WRITE based on the emission rate. Due to the deterministic feature of this schedule, the model can be easily implemented and trained. On the downside, anticipation from missing contents often fails to predict correct target tokens, and a fixed schedule could impede the model from speeding up or slowing down flexibly for source inputs. There are several works of SimulMT with many variants of the Wait-$k$ approach. For example, \cite{caglayan-etal-2020-simultaneous} explores whether additional visual context can complement missing source information. Furthermore, in \cite{zheng-etal-2020-opportunistic}, the opportunistic decoding technique is introduced which allows partial (certain length of suffix) corrections in a timely fashion. Finally, \cite{zheng-etal-2020-simultaneous} extended the wait-$k$ to an adaptive one by composing a set of fixed policies heuristically.

Upon the proposal by \cite{cho2016can}, various adaptive policies have been suggested by several works including \cite{gu-etal-2017-learning, zheng-etal-2019-simpler,zheng-etal-2019-simultaneous, arivazhagan-etal-2019-monotonic, Ma2020Monotonic}.
SimulMT proposed by \cite{cho2016can} use greedy decoding with heuristic waiting criteria to decide whether the model should read or emit, while \cite{gu-etal-2017-learning} utilize a pre-trained model with a reinforcement learning agent that maximizes quality and minimizes latency. Advancing this work, \cite{alinejad-etal-2018-prediction} proposes to add a new action PREDICT that anticipate future source words.
Recently, \cite{arivazhagan-etal-2019-monotonic} use hard attention to schedule the policy and introduced new differentiable average lagging metrics which can be integrated into training losses, and \cite{Ma2020Monotonic} incorporate this work into the multi-headed Transformer model. Furthermore, \cite{zhang-etal-2020-learning-adaptive} proposes an adaptive policy which learns to segment source input considering possible target output. Other researches including \cite{zheng-etal-2019-simpler} use separately trained oracles in the supervision of extracted action sequence.

\paragraph{Paraphrase:}

Translation is one of more common approaches for paraphrase generation. \citet{mallinson-etal-2017-paraphrasing} explore pivoting (translating a source sequence to a pivot language, then to a target language) to generate paraphrases and assess correlation between original and paraphrased sentences. Back-translation has also been explored for paraphrase generation \cite{wieting-etal-2017-learning, iyyer-etal-2018-adversarial}. Other techniques, such as translating with oversampling strategy have also bee studied \cite{chada-2020-simultaneous}.

On the other hand, various NMT research employ paraphrased data to overcome data limitation. \citet{edunov-etal-2018-understanding} show that source-paraphrased corpus generated with back-translation can significantly improve BLEU scores in NMT tasks. Similarly, \citet{khayrallah-etal-2020-simulated} directly implements paraphrasers in NMT training to improve translation quality.

\end{document}